\documentclass[10pt,twocolumn,letterpaper]{article}

\usepackage[pagenumbers]{cvpr} 

\usepackage[utf8]{inputenc} 
\usepackage[T1]{fontenc}    
\usepackage{url}            
\usepackage{booktabs}       
\usepackage{amsfonts}       
\usepackage{nicefrac}       
\usepackage{microtype}      
\usepackage{xcolor}         
\usepackage{graphicx}
\usepackage{caption} 
\usepackage{algorithm}
\usepackage[noend]{algpseudocode}
\usepackage{amsmath,amssymb}
\usepackage{array,multirow}
\usepackage{float}
\usepackage{wrapfig}
\usepackage{subcaption}
\def\etal{\emph{et al}.}

\newcommand{\bo}{\mathbf{o}}
\newcommand{\bs}{\mathbf{s}}
\newcommand{\ba}{\mathbf{a}}
\newcommand{\br}{\mathbf{r}}
\newcommand{\bg}{\mathbf{g}}

\newcommand{\st}{\bs_t}
\newcommand{\at}{\ba_t}
\newcommand{\gt}{\bg_t}

\newcommand{\stp}{\bs_{t+1}}

\newcommand{\states}{\mathcal{S}}
\newcommand{\actions}{\mathcal{A}}

\newcommand{\qattn}{Q}

\newcommand{\qattnp}{\theta}
\newcommand{\qrotgripp}{\phi}

\newcommand{\ctr}{f}

\newcommand{\obs}{\mathbf{o}}
\newcommand{\obst}{\obs_t}
\newcommand{\obstp}{\obs_{t+1}}

\newcommand{\rgb}{\mathbf{b}}

\newcommand{\pcd}{\mathbf{p}}
\newcommand{\proprio}{\mathbf{z}}

\newcommand{\replay}{\mathcal{D}}

\DeclareMathOperator{\E}{\mathbb{E}}

\DeclareMathOperator*{\argmax}{arg\,max}

\DeclareMathOperator*{\argmaxtwod}{\argmax2D}
\DeclareMathOperator*{\argmaxthreed}{\argmax3D}
\DeclareMathOperator*{\maxtwod}{\max2D}

\newcommand{\qdepthN}{N}
\newcommand{\qdepth}{n}
\newcommand{\vox}{\mathbf{v}}
\newcommand{\voxf}{\mathbf{V}}
\newcommand{\centre}{\mathbf{c}}
\newcommand{\res}{\mathbf{e}}

\usepackage[pagebackref,breaklinks,colorlinks]{hyperref}

\usepackage[capitalize]{cleveref}
\crefname{section}{Sec.}{Secs.}
\Crefname{section}{Section}{Sections}
\Crefname{table}{Table}{Tables}
\crefname{table}{Tab.}{Tabs.}

\def\confName{CVPR}
\def\confYear{2022}

\begin{document}

\title{Coarse-to-Fine Q-attention: Efficient Learning for \\Visual Robotic Manipulation via Discretisation
}

\author{Stephen James, Kentaro Wada, Tristan Laidlow, Andrew J. Davison\\
Dyson Robotics Lab\\
Imperial College London\\
{\tt\small \{slj12, k.wada18, t.laidlow15, a.davison\}@imperial.ac.uk}
}
\maketitle

\begin{abstract}
We present a coarse-to-fine discretisation method that enables the use of discrete reinforcement learning approaches in place of unstable and data-inefficient actor-critic methods in continuous robotics domains. This approach builds on the recently released ARM algorithm, which replaces the continuous next-best pose agent with a discrete one, with coarse-to-fine Q-attention. Given a voxelised scene, coarse-to-fine Q-attention learns what part of the scene to `zoom' into. When this `zooming' behaviour is applied iteratively, it results in a near-lossless discretisation of the translation space, and allows the use of a discrete action, deep Q-learning method. We show that our new coarse-to-fine algorithm achieves state-of-the-art performance on several difficult sparsely rewarded RLBench vision-based robotics tasks, and can train real-world policies, tabula rasa, in a matter of minutes, with as little as 3 demonstrations.
\end{abstract}
\vspace{-1em}

\section{Introduction}

In this paper, we are interested in a general real-world manipulation algorithm that can use a small number of demonstrations, along with a small amount of sparsely-rewarded exploration data, to accomplish a diverse set of tasks, both in simulation and the real world. To develop such an approach, two paradigms come to mind: imitation learning (IL) and reinforcement learning (RL). Imitation learning methods, such as behaviour cloning, suffer from compounding error due to covariate shift, while reinforcement learning suffers from long training times that often require millions of environment interactions. Recently however, Q-attention and the ARM system~\cite{james2021attention} has been shown to bypass many flaws that come with reinforcement learning, most notably the large training burden and exploration difficulty with sparsely-rewarded and long-horizon tasks.

\begin{figure}
\centering
\includegraphics[width=1.0\linewidth]{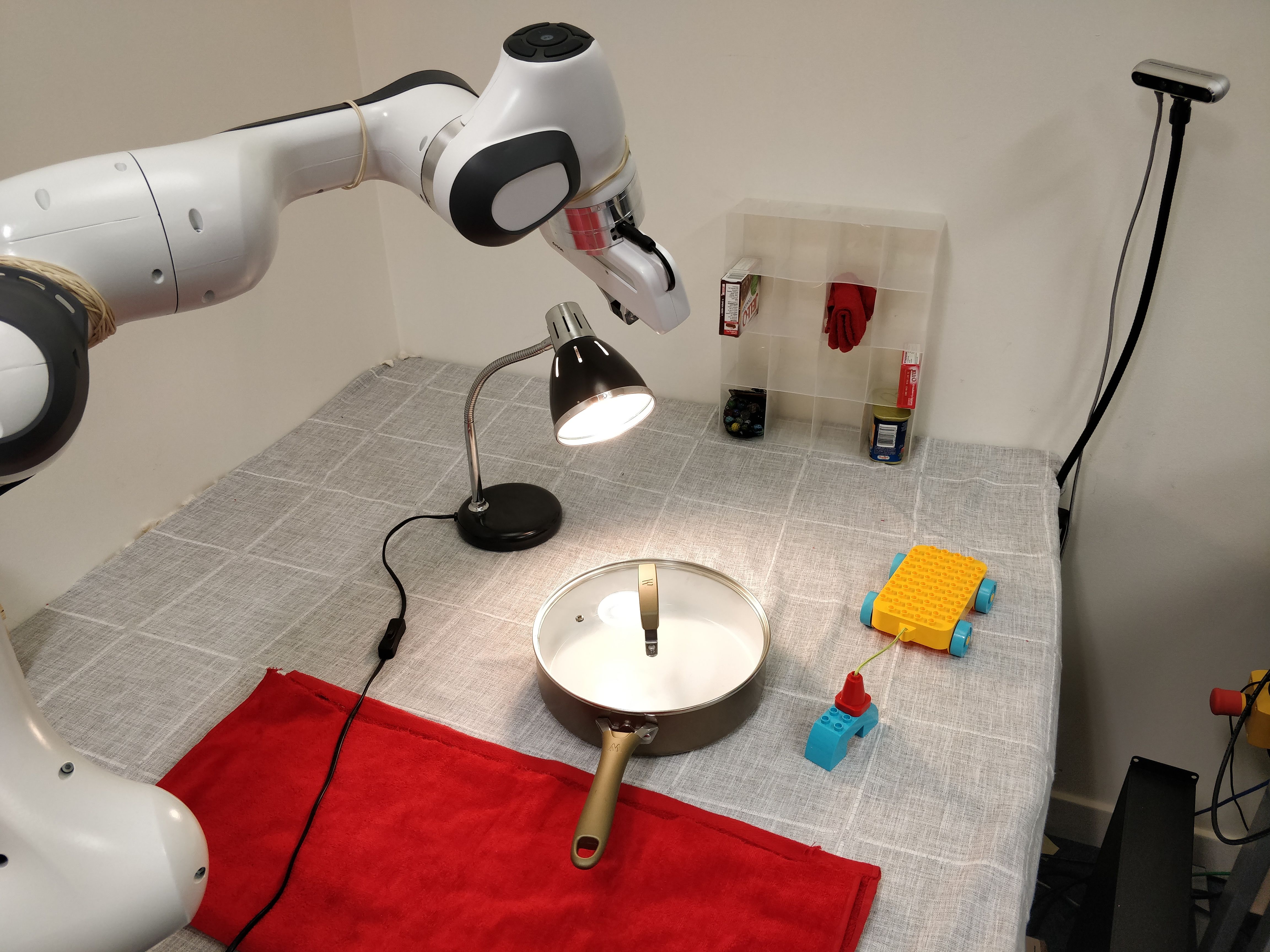}
\caption{C2F-ARM learns sparsely-rewarded tasks with only 3 demonstrations. Real-world tasks include: turning on a light, pulling cloth from shelf, pulling a toy car, taking a lid off a saucepan, and folding a towel.}
\label{fig:front_summary}
\end{figure}

Unfortunately, like many modern continuous control RL algorithms, ARM's next-best pose agent follows an actor-critic paradigm, which can be particularly unstable when learning from sparsely-rewarded and image-based tasks~\cite{james2021attention}: two properties that are particularly important for robot manipulation tasks. In this paper, we re-examine how best to represent the continuous control actions needed for robot manipulation, abandoning the standard actor-critic approach, in favour of a more stable discrete action Q-learning approach. The challenge therefore becomes how to effectively discretise 6D poses. Discretisation of rotation and gripper action is trivial given its bounded nature, but translation remains challenging given that is usually a much larger space. We solve this problem via a coarse-to-fine Q-attention, where we start with a coarse voxelisation of the translation space, use 3D Q-attention to identify the next most interesting point, and gradually make the resolution higher at each point.

\begin{figure*}
\centering
\includegraphics[width=1.0\linewidth]{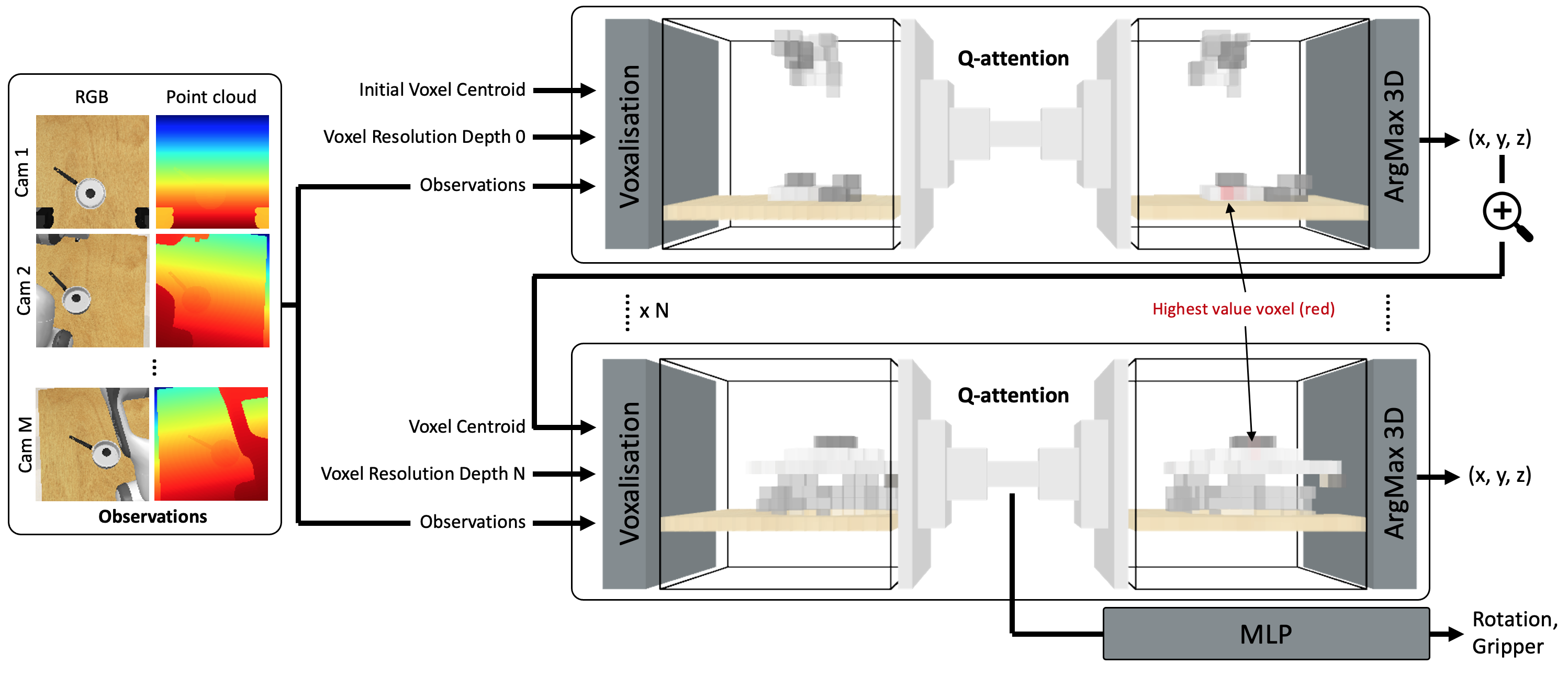}
\caption{Summary of coarse-to-fine Q-attention. Observation data (RGB and point cloud) from $M$ cameras are given to each depth of the Q-attention. Each depth of the Q-attention gives the locations of the most interesting point in space (at the current resolution), which is then used as the voxel centroid for the Q-attention at the next depth. Intuitively, this can be thought of as `zooming' into a specific part of the scene to gain more accurate 3D information. The highlighted red voxel corresponds to the highest value.}
\label{fig:qattention_summary}
\vspace{-1em}
\end{figure*}

With this new coarse-to-fine Q-attention, we present our Coarse-to-Fine Attention-driven Robotic Manipulation (C2F-ARM) system. We benchmark the system in simulation against other robot learning algorithms from both the reinforcement learning and imitation learning literature, and show that C2F-ARM is more sample-efficient and stable to train than other methods. We also show that C2F-ARM is capable of learning 5 diverse sets of sparsely-rewarded real-world tasks from only 3 demonstrations. 

To summarise, the paper presents the following three contributions: \textbf{(1)} A novel way to discretise the translation space via coarse-to-fine Q-attention, allowing us to discard the often unstable actor-critic framework for a simpler deep Q-learning approach. \textbf{(2)} Our manipulation system, C2F-ARM, which uses the coarse-to-fine Q-attention along with a control agent to achieve sample-efficient learning of sparsely-rewarded tasks in both simulation and real-world. \textbf{(3)} The first use of a voxel representation for vision-based reinforcement learning for 6D robot manipulation. Code in supplementary material, and videos found at: \url{sites.google.com/view/c2f-q-attention}.

\section{Related Work}

\textbf{Learning for manipulation}. Recent trends in learning for manipulation have seen continuous-control reinforcement learning algorithms, such as PPO~\cite{schulman2017proximal}, DDPG~\cite{lillicrap2015continuous}, TD3~\cite{fujimoto2018addressing}, and SAC~\cite{haarnoja2018soft}, trained on a variety of tasks, including cloth manipulation~\cite{matas2018sim}, lego stacking~\cite{haarnoja2018composable}, pushing~\cite{pinto2017asymmetric}, and in-hand manipulation~\cite{rajeswaran2017learning}. These approaches rely on the actor-critic formulation, which is often sample-inefficient and unstable to train. One effort to decrease instability involves using alternative policy parameterisations over Gaussian ones, e.g., the Bignham distribution for a next-best pose action~\cite{james2022bingham}. Alternatively, discretisation results in a significant reduction to the action space, as well as the use of simpler approaches, such as Q-learning. Our work is not the first use of discrete actions for visual manipulation; James \etal~\cite{james20163d} discretised the joint space, where in each step the agent could choose to move one of the 6 joints by 1 degree. An alternative to discretising the joint space, is to discretise the planar workspace, where pixels from a top-down camera act as high-level actions, such as grasping~\cite{morrison2018closing}, pushing~\cite{zeng2018learning}, and pick-and-place~\cite{zeng2020transporter}. However, it is unclear how these can extend beyond top-down pick-and-place tasks, such as some of the ones featured in this paper, e.g. stacking wine and taking an object from a shelve. Our paper presents a full 6D manipulation system that can extend to a range of tasks, not just top-down ones. The term `coarse-to-fine' has recently been used in the context of grasp detection~\cite{wu2019pixel}, but with a different meaning, where the coarse part is a grasp confidence grid, and the fine part refers to a refinement stage of the coarse prediction. Our work on the other hand uses coarse-to-fine to refer to the idea that our Q-attention `zooms' into a specific part of the scene to gain more accurate 3D information. Related to our coarse-to-fine mechanism is the work of Gualtieri \etal~\cite{gualtieri2018learning,gualtieri2020learning}, which takes in a point-cloud (in heightmap form) and gradually zooms into a local region to generate a grasp and place location; however, this work is restricted `pick-and-place' tasks, while our method is a \textit{general} 6D manipulation algorithm where the arm has full autonomy to move anywhere in the scene, with no pre-programmed `grasp' and `place' motions. This results in a system that can be applied to a range of tasks without modification, as is evident by our RLBench and real-world experiments.

\textbf{Voxel representation for manipulation}. Modelling 3D environments via voxels dates back as far as the 1980s~\cite{roth1989building,moravec1996robot}. Since a voxel grid can store arbitrary values in each voxel, prior work has used it for various representations (geometry, semantics, learned features) for navigation and manipulation. Most works that explore voxel representations use them for navigation~\cite{hornung2010humanoid,dryanovski2010multi,hornung2012navigation}; however, there are some notable exceptions from the manipulation domain. Wada \etal~\cite{wada2017probabilistic} used a voxel grid to store occupancy and semantics of objects in a cluttered scene to select the next target object and grasp point.  MoreFusion~\cite{wada2020morefusion} is a system that uses voxels to perform multi-object reasoning to improve 6D pose estimation; the system gives accurate object poses and can perform precise pick-and-place in cluttered scenes.  Recently, it has become common to use a voxel representation with a learning-based model. Song \etal~\cite{song2020grasping} and Breyer \etal~\cite{breyer2021volumetric} fed a voxel grid representation to a neural network to generate 6DoF grasp actions; these works differ to ours in that they consider grasping in a supervised learning domain, where as our system falls within the realm of reinforcement learning with a full 6D action space, where grasping is but only one component. Moreover, reinforcement learning brings with it many challenges that are not present in supervised learning, e.g. exploration, sparse rewards, and long-horizon planning. The only work we are aware of that uses voxels with RL is in a task to find a red cube in clutter~\cite{novkovic2020object}; however the task assumes access to both and object detector and planar workspace, and the voxel grid is not processed directly by the RL agent, but instead is cropped and flattened to a small 68 dimensional vector. Our coarse-to-fine voxelisation allows us to directly process the 3D voxel grid of the whole scene, and does not assume access to a planar workspace or an object detector.

\section{Background}

\subsection{Reinforcement Learning}

The reinforcement learning paradigm assumes an agent interacting with an environment consisting of states $\bs \in \states$, actions $\ba \in \actions$, and a reward function $R(\st,\at)$, where $\st$ and $\at$ are the state and action at time step $t$ respectively. The goal of the agent is then to discover a policy $\pi$ that results in maximising the expectation of the sum of discounted rewards: $\E_\pi [\sum_t \gamma^t R(\st, \at)]$, where future rewards are weighted with respect to the discount factor $\gamma \in [0, 1)$. Each policy $\pi$ has a corresponding value function $Q(s, a)$, which represents the expected return when following the policy after taking action $\ba$ in state $\bs$.

The Q-attention module~\cite{james2021attention} (discussed in Section \ref{sec:back:arm}) builds from Deep Q-learning \cite{mnih2015human}; a method that approximates the value function $Q_\psi$, with a deep convolutional network, whose parameters $\psi$ are optimised by sampling mini-batches from a replay buffer $\replay$ and using stochastic gradient descent to minimise the loss: $\E_{(\st, \at, \stp) \sim \replay} [ (\br + \gamma \max_{\ba'}Q_{\psi'}(\stp, \ba') - Q_{\psi}(\st, \at))^2]$, where $Q_{\psi'}$ is a target network; a periodic copy of the online network $Q_\psi$ which is not directly optimised.

\subsection{Attention-driven Robot Manipulation (ARM)}
\label{sec:back:arm}

ARM~\cite{james2021attention} introduced several core concepts that facilitate the learning of vision-based robot manipulation tasks. These included Q-attention, keypoint detection, demo augmentation, and a high-level next-best pose action space. Most notable of these is the Q-attention, which is used in this work to discretise the large translation space. We briefly outline Q-attention below.

Given an observation, $\obs$ (consisting of an RGB image, $\rgb$, an organised point cloud, $\pcd$, and proprioceptive data, $\proprio$), the Q-attention module, $\qattn_\qattnp$, outputs 2D pixel locations of the next area of interest. This is done by extracting the coordinates of pixels with the highest value: $(x, y) = \argmaxtwod_{\ba'} \qattn_{\qattnp}(\obs, \ba')$, where $\argmaxtwod$ is an \textit{argmax} taken across two dimensions. These pixel locations are used to crop the RGB image and organised point cloud inputs and thus drastically reduce the input size to the next stage of the pipeline; this next stage is an actor-critic next-best pose agent. The parameters of the Q-attention are optimised by using stochastic gradient descent to minimise the loss:
\begin{multline}
\label{eq:qattention_2d}
J_{\qattn}(\qattnp) = \E_{(\obst, \at, \obstp) \sim \replay} [ (\br + \gamma \maxtwod_{\ba'}\qattn_{\qattnp'}(\obstp, \ba') \\ - \qattn_{\qattnp}(\obst, \at))^2 + \lVert \qattn \rVert],
\end{multline}
where $\qattn_{\qattnp'}$ is the target Q-function, and $\lVert \qattn \rVert$ is the \textit{Q regularisation} --- an $L2$ loss on the per-pixel output of the Q function.

Keyframe discovery and demo augmentation were another two important techniques introduced in ARM~\cite{james2021attention}. Rather than simply inserting demonstrations directly into the replay buffer, the keyframe discovery strategy chooses interesting keyframes along demonstration trajectories that are fundamental to training the Q-attention module. Demo augmentation stores the transition from intermediate points along a trajectory to the keyframe states, rather than storing the transition from an initial state to a keyframe state. This greatly increases the amount of initial demo transitions in the replay buffer. This is not specific to Q-attention, and is applied to all methods (including baselines) in this work. Note that keyframe discovery is crucial for training our coarse-to-fine Q-attention agent, as the keyframes act as explicit supervision to help guide the Q-attention to choose relevant areas to `zoom' into during the initial phase of training.

\section{Method}

\begin{algorithm*}[tb]
\caption{Coarse-to-Fine Attention-driven Robot Manipulation (C2F-ARM)}
\label{alg:arm}
\begin{algorithmic}
    \State Initialise the $\qdepthN$ Q-attention networks $\qattn_{\qattnp_1}, \dots, \qattn_{\qattnp_\qdepthN}$ with random parameters $\qattnp_1, \dots, \qattnp_\qdepthN$.
    \State Initialise the rotation \& gripper Q network $\qattn_{\qrotgripp}$ with random parameters $\qrotgripp \subset \qattnp_\qdepthN$.
    \State Initialise target networks $\qattnp'_1 \leftarrow \qattnp_1, \dots, \qattnp'_\qdepthN \leftarrow \qattnp_\qdepthN$.
    \State Initialise replay buffer $\replay$ with demos and apply keyframe selection and demo augmentation
    \For{each iteration}
	    \For{each environment step $t$}
	        \State $\obs_t \leftarrow (\rgb_t, \pcd_t, \proprio_t)$
	        
	        \State $\centre^0 \leftarrow$ Scene centroid
	        \State $\text{coords} \leftarrow [\text{ }]$  \Comment{List to keep coords of each Q-attention depth}

	        \For{each ($\qdepth$ of $\qdepthN$) Q-attention depths}
	        
	            \State $\vox^\qdepth \leftarrow \voxf(\obs_t, \res^\qdepth, \centre^\qdepth)$ \Comment{Voxelise with given resolution \& centroid}
	            
    	        \State $(x^\qdepth, y^\qdepth, z^\qdepth) \leftarrow \argmaxthreed_{\ba'} \qattn_{\qattnp_{\qdepth}}(\vox^\qdepth, \ba')$ \Comment{Use Q-attention to get voxel coords}
    	        \State $\text{coords.append(} (x^\qdepth, y^\qdepth, z^\qdepth) \text{)}$
    	        
    	       \If{$\qdepth == N$}
    	            \State $\alpha, \beta, \gamma, \omega \leftarrow \argmax_{\ba^h} \qattn^{h}_{\qrotgripp}(\tilde{\vox^\qdepthN}, \ba^h)$ for $h \in \{0, 1, 2, 3\}$ \Comment{Rotation \& gripper from bottleneck features $\tilde{\vox^\qdepthN}$}
    	       \EndIf
    	        
    	        \State $\centre^{\qdepth+1} \leftarrow (x^\qdepth, y^\qdepth, z^\qdepth)$  \Comment{Voxel coords give centroid of next Q-attention depth}
	        \EndFor
	        
	    \State $\at \leftarrow (\centre^{\qdepthN}, \alpha, \beta, \gamma, \omega)$ \Comment{The next-best pose}
	    \State $\obs_{t+1}, \br \leftarrow env.step(\at)$  \Comment{Use motion planning to bring us to the next-best pose.}
        \State $\mathcal{D} \leftarrow \replay \cup \left\{(\obs_t, \at, \br, \obs_{t+1}, \text{coords})\right\}$ \Comment{Store the transition in the replay buffer}
	        
	    \EndFor
    	\For{each gradient step}
    	    \State $\qattnp_\qdepth \leftarrow \qattnp_\qdepth - \lambda_{\qattn} \hat \nabla_{\qattnp_\qdepth} J_{\qattn}(\qattnp_\qdepth)$ for $\qdepth \in \{0, \dots, \qdepthN\}$ \Comment{Update parameters}
    	    \State $\qattnp'_\qdepth \leftarrow \tau \qattnp_\qdepth + (1-\tau) \qattnp'_\qdepth$ for $\qdepth \in\{0, \dots, \qdepthN\}$ \Comment{Update target network weights}
    	\EndFor
    \EndFor
\end{algorithmic}
\end{algorithm*}

Our system, C2F-ARM (Algorithm \ref{alg:arm}), can be split into 2 core phases. Phase 1 (Section \ref{sec:phase1}) consists of the coarse-to-fine 3D Q-attention agent, which starts by voxelising the entire scene in a coarse manner, and then recursively makes the resolution finer until we are able to extract a continuous 6D next-best pose. Phase 2 (Section \ref{sec:phase2}) is a low-level control agent that accepts the predicted next-best pose and executes a series of actions to reach the given goal pose. Before training, we fill the replay buffer with demonstrations using keyframe discovery and demo augmentation~\cite{james2021attention}.

The system assumes we are operating in a partially observable Markov decision process (POMDP), where an observation $\obs$ consists of an RGB image, $\rgb$, an organised point cloud, $\pcd$, and proprioceptive data, $\proprio$. Actions consist of a 6D (next-best) pose and gripper action, and the reward function is sparse, giving $100$ on task completion, and $0$ for all other transitions.

\subsection{Coarse-to-fine Q-attention}
\label{sec:phase1}

The key contribution of this paper is the discretisation of the translation state space via the Q-attention, thereby allowing discrete-action reinforcement learning algorithms to be used for recovering continuous actions. Our method formulates the translation prediction as a series of coarse-to-fine deep Q-networks which accepts voxelised point cloud features, and outputs per-voxel Q-values. The highest-valued voxel represents the next-best voxel, whose location is used as the centre of a higher-resolution voxelisation in the next step. Note that `higher-resolution' could be interpreted in one of 3 ways: (1) keeping the volume the same but increasing the number of voxels, (2) keeping the number of voxels the same but reducing the volume, or (3) a combination of both. We opt for (2), as this gives us the higher resolution, while keeping the memory footprint low. Intuitively, this coarse-to-fine Q-attention can be thought of as `zooming' into a specific part of the scene to gain more accurate 3D information, whereas formally, each Q-attention agent operates at different resolutions to view the same underlying POMDP. The coarse-to-fine prediction is applied several times, which gives near-lossless prediction of the continuous translation. Rotation and gripper action prediction is simpler due to its bounded nature; these are predicted in the final depth of the Q-attention as an additional branch. The coarse-to-fine Q-attention is summarised in Figure \ref{fig:qattention_summary}.

Formally, we define a voxelisation function $\vox^\qdepth = \voxf(\bo, \res^\qdepth, \centre^\qdepth)$, which takes the observation $\obs$, a voxel resolution $\res^\qdepth$, and a voxel grid centre $\centre^\qdepth$, and returns a voxel grid $\vox^{\qdepth} \in \mathbb{R}^{x×y×z×(3+M+1)}$ at depth $\qdepth$, where $\qdepth$ is the depth/iteration of the Q-attention, and where each voxel contains the 3D coordinates, $M$ features (e.g. RGB values, features, etc), and an occupancy flag.

Given our Q-attention function $\qattn_{\qattnp_{\qdepth}}$ at depth $\qdepth$, we extract the indicies of the voxel with the highest value:
\begin{equation}
\label{eq:extractxyz}
\vox_{ijk}^{\qdepth} = \argmaxthreed_{\ba'} \qattn_{\qattnp_{\qdepth}}(\vox^\qdepth, \ba'),
\end{equation}
where $\vox_{ijk}$ is the extracted voxel index located at $(i, j, k)$, and $\argmaxthreed$ is an \textit{argmax} taken across three dimensions (depth, height, and width).

By offsetting the centre of the current voxelisation with the extracted indicies, we can trivially extract the $(x^\qdepth, y^\qdepth, z^\qdepth)$ location of that voxel. For ease of readability, we henceforth assume that $\argmaxthreed$ also performs the conversion to world coordinates, to directly give $(x^\qdepth, y^\qdepth, z^\qdepth)$. As the extracted coordinates represent the next-best coordinate to voxelise at a higher resolution, we set these coordinates to be the voxel grid centre $\centre$ for the next depth: $\centre^{\qdepth+1} = (x^\qdepth, y^\qdepth, z^\qdepth)$. However, if this is the last depth of the Q-attention, then $\centre^{\qdepthN} = \centre^{\qdepth+1}$ represents the continuous representation of the translation (i.e. the translation component of the next-best pose agent).

Due to the fact that the rotation space and gripper space is much smaller than the translation, we can resort to a much simpler Q-value prediction. The rotation of each axis is discretised in increments of $5$ degrees, while the gripper is discretised to be either open or closed. These components are recovered from an MLP branch (with parameters $\qrotgripp$) of the final Q-attention depth:
\begin{equation}
\label{eq:rotgrp}
\alpha, \beta, \gamma, \omega \leftarrow \argmax_{\ba^h} \qattn^{h}_{\qrotgripp}(\tilde{\vox^\qdepthN}, \ba^h) \text{ for } h \in \{0, 1, 2, 3\},
\end{equation}
where $\alpha, \beta, \gamma$ represent the individual rotation axis, $\omega$ is the gripper action, and $\tilde{\vox^\qdepthN}$ are the bottleneck features from the final Q-attention depth. We empirically found the discretisation of the rotation axis to be robust to a range of values from $1$ to $10$. The final action then becomes $\at = (\centre^{\qdepthN}, \alpha, \beta, \gamma, \omega)$.

The coarse-to-fine Q-attention shares the same motivation that was was laid out in ARM~\cite{james2021attention}, i.e. that our gaze focuses sequentially on objects being manipulated~\cite{land1999roles}, however, its role in the manipulation system is different. ARM~\cite{james2021attention} uses Q-attention to reduce the image resolution to a next-best pose phase (by cropping $128 \times 128$ observations to $16 \times 16$), while the role of coarse-to-fine Q-attention is to discretise the otherwise large translation space. This highlights the versatility of Q-attention.

\subsection{Control Agent}
\label{sec:phase2}

The control agent remains largely unchanged from ARM~\cite{james2021attention}. Given the next-best pose output from the previous stage, we give this to a goal-conditioned control function $\ctr(\st, \gt)$, which given state $\st$ and goal $\gt$, outputs motor velocities that moves the end-effector towards the goal. This function can take on many forms, but two noteworthy solutions would be either motion planning in combination with a feedback-control or a learnable policy trained with imitation/reinforcement learning. Given that the environmental dynamics are limited in the benchmark, we opted for the motion planning solution.

Given the target pose, we perform path planning using the SBL~\cite{sanchez2003single} planner within OMPL~\cite{sucan2012ompl}, and use `Reflexxes Motion Library' for on-line trajectory generation. If the target pose is out of reach, we terminate the episode and supply a reward of $-1$. For the simulated experiments, the path planning and trajectory generation is conveniently encapsulated by the \textit{`EndEffectorPoseViaPlanning'} action mode in RLBench \cite{james2019rlbench}, while for the real-world experiments, we use ROS to handle the planning and trajectory generation.

Once the agent reaches the end of the trajectory, the new observation is given $\obs_{t+1}$, and is stored in the replay buffer as experience for the coarse-to-fine Q-attention: $\mathcal{D} \leftarrow \replay \cup \left\{(\obs_t, \at, \br, \obs_{t+1}, \text{coords})\right\}$, where $\mathcal{D}$ is the replay buffer, and $\text{coords}$ is the list of extracted coordinates $(x^\qdepth, y^\qdepth, z^\qdepth) \text{ for } \qdepth \in \{0, \dots, \qdepthN\}$.

\begin{figure*}
\centering
\includegraphics[width=1.0\linewidth]{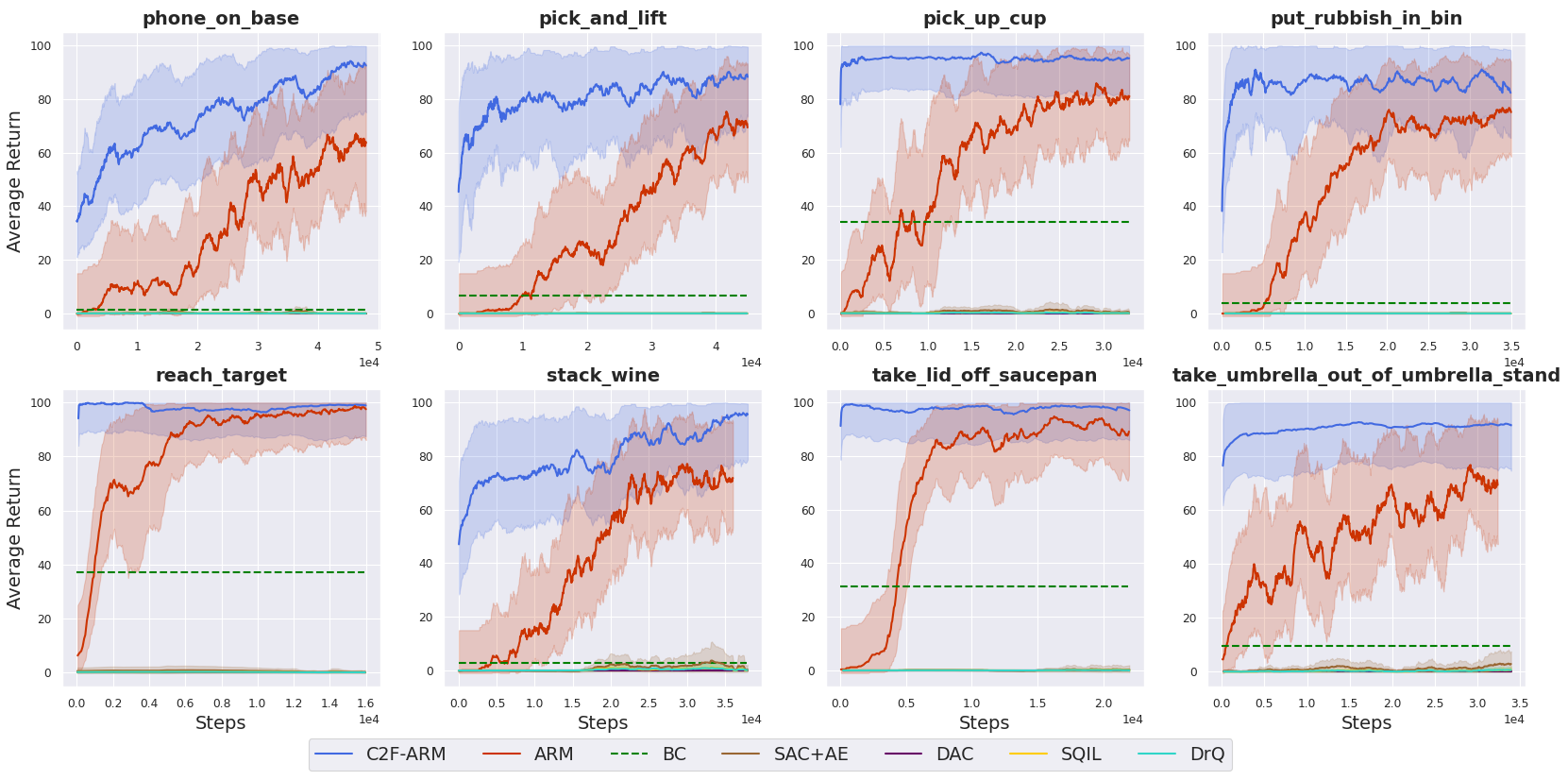}
\caption{Learning curves for 8 RLBench tasks. In addition to our method (C2F-ARM), we include the same baselines as in previous work: ARM~\cite{james2021attention}, BC, SAC+AE~\cite{yarats2019improving}, DAC~\cite{kostrikov2018discriminator} (an improved, off-policy version of GAIL~\cite{ho2016generative}), SQIL~\cite{reddy2019sqil}, and DrQ~\cite{kostrikov2020image}. To further show the sample efficiency of C2F-ARM, our method \textbf{only receive 10 demos}, while all other \textbf{baselines receiving 100 demos}, giving baseline methods a big advantage. Demos are stored in the replay buffer prior to training; giving baseline methods a big advantage. Solid lines represent the average evaluation over 5 seeds, where the shaded regions represent the standard deviation across those trials. Evaluation starts at step 100.}
\label{fig:main_baseline_results}
\end{figure*}

\subsection{Network Architecture}

Each Q-attention layer follows a light-weight U-Net style architecture \cite{ronneberger2015u}, but uses 3D convolutions rather than 2D ones. Our U-Net encoder features 3 Inception-style blocks~\cite{szegedy2015going}, with 64 input-output channels, and a $3\times3$ maxpool after each block, while our U-Net decoder features 3 Inception-Upsample-Inception blocks. Note that the final Q-attention layer is also used for the rotation and gripper prediction; this is achieved by concatenating the maxpooled and soft-argmax values after each of the Inception blocks in the decoder, and sending them through 2 fully connected layers each with 256 nodes. Finally, these features are passed through a final fully-connected layer which gives the rotation and gripper discretisation. The initial voxel centroid was set to be the centre of the scene. Voxelisation code has been adapted from Ivy~\cite{lenton2021ivy}.

\section{Results}

The results can be broken into three core sections: (1) simulation results using RLBench~\cite{james2019rlbench} to benchmark our algorithm against other popular robot learning algorithms using only the front-facing camera. (2) additional simulation results in RLBench where we evaluate our method on additional tasks and perform an ablation study into the robustness of the coarse-to-fine approach. (3) Real-world results where we show that the sample-efficiency in simulation is also present when training from scratch in the real world.

\subsection{Simulation}

For our simulation experiments, we use RLBench~\cite{james2019rlbench}. RLBench was chosen due to its emphasis on vision-based manipulation benchmarking and because it gives access to a wide variety of tasks with demonstrations. Each task has a completely sparse reward of $100$ which is given only on task completion, and $0$ otherwise.
In the following set of experiments, unless otherwise stated, our method uses a coarse-to-fine depth of $2$, each with a voxel grid size of $16^3$. All methods get the same observations (RGB and point clouds), and have the same action space (next-best pose).

\subsubsection{Comparison to other Robot Learning Methods}

\begin{figure*}
\centering
\includegraphics[width=1.0\linewidth]{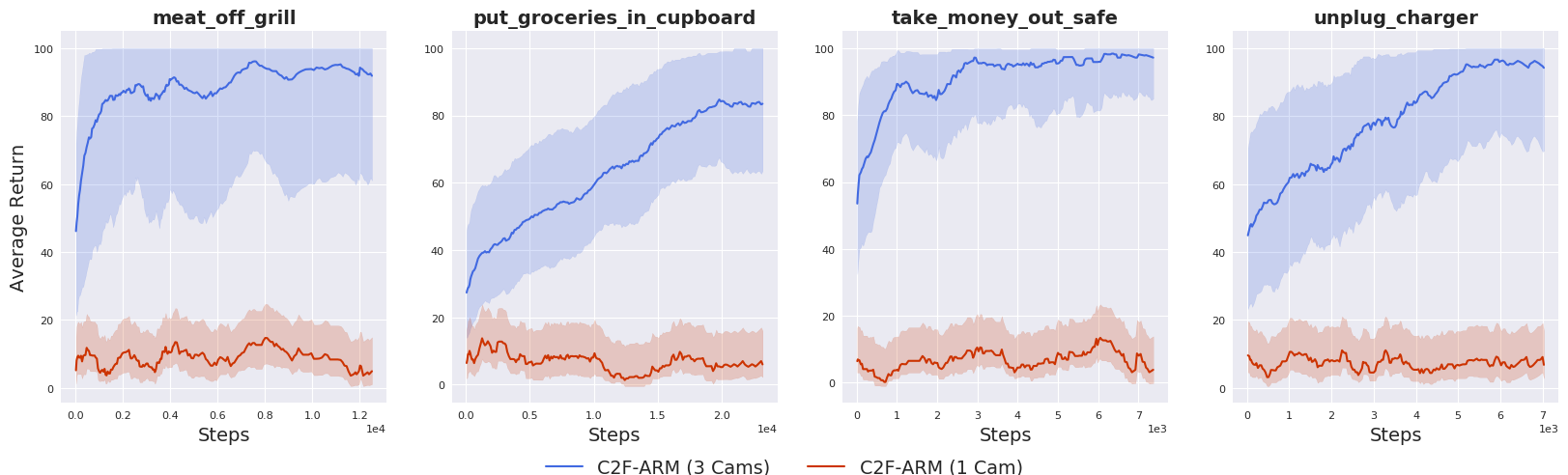}
\caption{Learning curves for 4 additional RLBench tasks that are difficult or impossible to achieve with only the front-facing camera. The 3 cameras used are the wrist, left shoulder, and right shoulder. Solid lines represent the average evaluation over 5 seeds, where the shaded regions represent the standard deviation across those trials.}
\label{fig:ablations_multi_cam}
\end{figure*}
\begin{figure}
\centering
\includegraphics[width=1.0\linewidth]{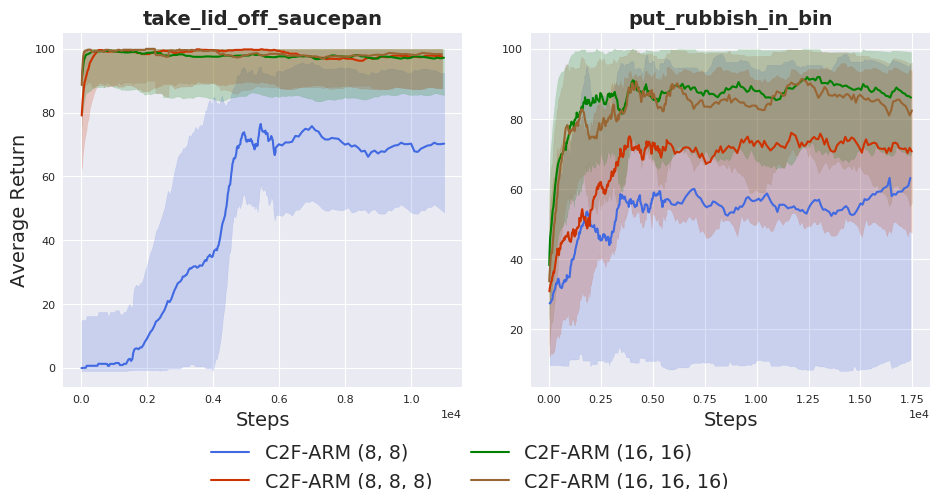}
\caption{Investigation of the Q-attention depth and voxel volume across the \textit{`take\_lid\_off\_saucepan'} and \textit{`put\_rubbish\_in\_bin'} tasks. The numbers in the brackets indicate the coarse-to-fine depth along with the voxel grid size; e.g. $(8, 8)$ represents a coarse-to-fine depth of 2, each with voxel grid size of $8^3$.}
\label{fig:ablations_voxel_size}
\end{figure}

To compare to other robot learning methods, we select the same 8 tasks as in James \etal~\cite{james2021attention}; these are task that are achievable from using only the front-facing camera. Figure \ref{fig:main_baseline_results} shows the results of this comparison. We selected a range of common baselines from the imitation learning and reinforcement learning literature; these include: include ARM~\cite{james2021attention}, behavioural cloning (BC), SAC+AE~\cite{yarats2019improving}, DAC~\cite{kostrikov2018discriminator} (an improved, off-policy version of GAIL~\cite{ho2016generative}), SQIL~\cite{reddy2019sqil}, and DrQ~\cite{kostrikov2020image}. Each of these baselines outputs a 7D action; 6D for the pose, and 1D for the gripper state. The control agent (as described in Section \ref{sec:phase2}) is then used to take the arm to the outputted pose. This means that all baselines act as an alternative to the coarse-to-fine Q-attention. Baselines use the same architectures as presented in James \etal~\cite{james2021attention}.

All methods (C2F-ARM, ARM, BC, SAC+AE, DAC, SQIL, and DrQ) feature keyframe discovery and demo augmentation~\cite{james2021attention}, and receive the same demonstration sequences, which are loaded into the replay buffer prior to training. Note that C2F-ARM does not require as many demonstrations as the other methods (as is evident in Section \ref{sec:realworld}), and so is given only 10 demos, while baselines receive 100 (giving only 10 demos to baselines made performance significantly worse).

The results in Figure \ref{fig:main_baseline_results} show that our method outperforms ARM\cite{james2021attention} by a large margin; either by attaining an overall higher performance, or attaining the same performance but in substantially fewer environment steps. The particularly poor performance of conventional actor-critic methods highlights their instability in challenging vision-based, sparsely-rewarded tasks. We wish to stress that perhaps given enough training time some of these baseline methods may eventually start to
succeed, however we found no evidence of this. To get the reinforcement learning baselines to successfully train on these tasks, it would most likely need access to privileged simulation-only abilities (e.g. reset to demonstrations, asymmetric actor-critic, auxiliary tasks, or reward shaping); this would then render the approach impractical for real-world training. Real-world reward shaping in particular is very cumbersome; for example, shaping the reward for our real world ‘lifting saucepan lid’ task would first require us to build a lid tracking system, before reward design can even begin. Moreover, reward shaping is notoriously difficult to get right as the complexity of the task increases~\cite{rajeswaran2017learning}.

In terms of wall-clock time, the inference time is increased by a factor of $2$ due to the use of 3D convolutions over the baselines use of 2D convolutions; however the inference time is negligible compared to the time it takes for the arm to navigate to the next-best pose. We also note as a purely qualitative observation, that C2F-ARM required little to no hyperparameter tuning, while baselines required a substantial amount.

\subsubsection{Multi-camera and Ablations}

The second set of simulation experiments evaluates C2F-ARM with multiple cameras. One of the weaknesses of ARM~\cite{james2021attention} was its inability to trivially handle multi-camera environments; for this reason, tasks were chosen that could be done with only the front-facing camera.
However, for real-world robotics, it is unreasonable to expect that a single camera will always contain the information required to accomplish a task; in reality, robotic systems are required to fuse information from multiple cameras into a single representation. For this reason, Figure \ref{fig:ablations_multi_cam} shows an additional 4 tasks from RLBench which we believe to be difficult to accomplish by only using the front-facing camera. For each of these tasks, we run our method using 3 cameras (wrist, left shoulder, and right shoulder), and compare this to using only the front-facing camera. Because all cameras are fused into a single voxel grid, no part of the system needs to be modified when using additional cameras. The results in Figure \ref{fig:ablations_multi_cam} clearly show that these tasks cannot be done with only a single camera, and that C2F-ARM can perform well when given the appropriate camera information. 

\begin{figure*}
\centering
\includegraphics[width=1.0\linewidth]{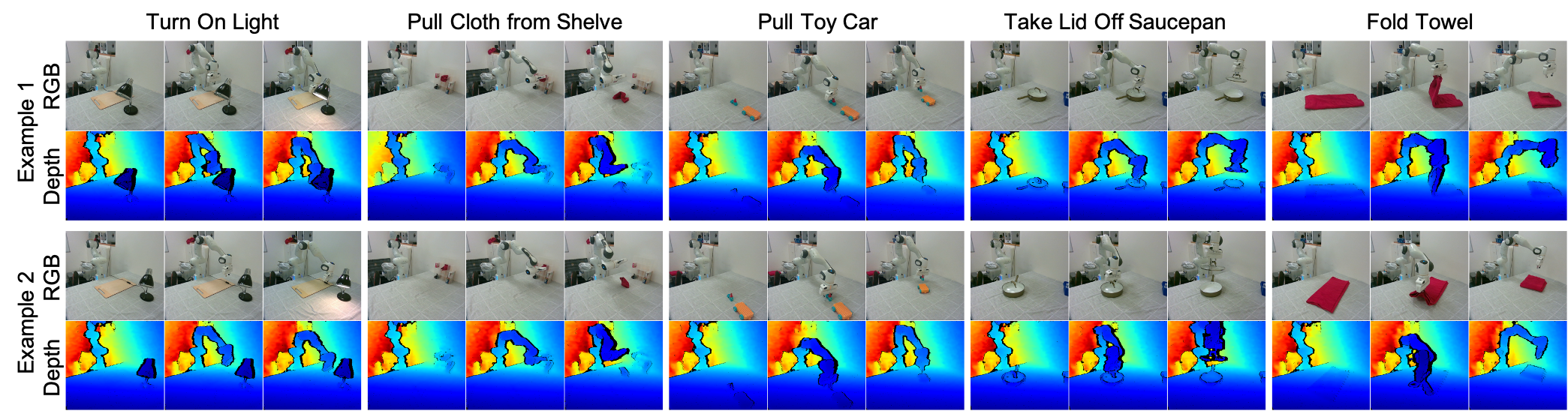}
\caption{Two examples of successful trials performed with C2F-ARM on the tasks: turn on light, pull cloth from shelf, pull toy car, take lid off saucepan, and fold towel. The agent only received 3 demonstrations. Each column for each tasks shows the RGB-D observations at $t=0$, $t=T/2$, and $t=T$.}
\label{fig:real_world_task_grid_text}
\vspace{-1em}
\end{figure*}

For the final set of simulation experiments, we evaluate how robust C2F-ARM is when altering the number of coarse-to-fine depth and the volume of the voxels. Figure \ref{fig:ablations_voxel_size} shows that our method is robust to a range of coarse-to-fine depths and voxel grid sizes, though note that as the voxel grid size and coarse-to-fine depth jointly decrease, performance begins to deteriorate. Note that the coarsest setup (8, 8) understandably performs the worst, as there are only two layers of a very coarse $8^3$ voxel grid, making the scene understanding phase difficult, particularly at the finest phase, where each voxel will contain many points. We hypothesise that voxelising image features (rather than raw RGB values) would perform better at these coarser setups; we leave this for future work.
Figure \ref{fig:ablations_voxel_size} also suggests that increasing the voxel grid size leads to better performance, though note that this will lead to a larger memory footprint. Note, that if memory is a limiting factor, then performance gains can be had by simply increasing the coarse-to-fine depth, with only a small increase to the memory footprint; i.e. the memory footprint of an additional depth (e.g. $(8,8) \rightarrow (8,8,8)$) is significantly less than the footprint of moving to a larger voxel grid size (e.g. $(8,8) \rightarrow (16,16)$).

\subsection{Real World}
\label{sec:realworld}
To further show the sample efficiency of our method, we train on 5 real-world tasks from scratch, which can be seen in Figure \ref{fig:real_world_task_grid_text}. At the beginning of each episode, the objects in the tasks are moved randomly within the robot workspace. We train each of the tasks until the agent achieves 4 consecutive successes. The approximate time to train each task are: pulling cloth from shelf ($\sim26$ minutes), pulling a toy car ($\sim18$ minutes), taking a lid off a saucepan ($\sim6$ minutes), folding a towel ($\sim24$ minutes), and turning on a light ($\sim42$ minutes). We use the Franka Emika Panda, and a single RGB-D RealSense camera. All tasks receive 3 demonstrations which are given through the HTC Vive VR system. These qualitative results are best seen via the full, uncut training video of each of the 5 tasks, located on the project website.

\section{Discussion and Conclusion}

We have presented Coarse-to-Fine Attention-driven Robot Manipulation (C2F-ARM), which is an algorithm that utilises a coarse-to-fine Q-attention and allows discretisation of the translation space. With this discretisation, we are able to diverge from unstable actor-critic methods and instead use a more stable deep Q-learning method. The result is a sample-efficient robot learning algorithm that outperforms others and can rapidly learn real-world tasks. 

C2F-ARM can be considered as an improved, discrete-action version of ARM~\cite{james2021attention}. There are 3 key differences to the original ARM system: (1) the role and architecture of Q-attention has changed; in ARM, the role of the 2D Q-attention was to act as a hard-attention that would give crops to the actor-critic next-best-pose agent, whereas in C2F-ARM the role of the 3D Q-attention is to be recursively applied in a coarse-to-fine manner in order to discretise the large translation space. (2) The number of stages in the system has decreased; ARM was a 3-stage system, consisting of Q-attention, next-best-pose agent, and the control agent, whereas C2F-ARM removes the need for the actor-critic next-best-pose agent, and so consists only of the coarse-to-fine Q-attention and control agent. (3) C2F-ARM seamlessly supports multiple cameras or a single moving camera; ARM was not suited for multiple cameras, due to the undefined behaviour when a camera observation did not feature any interesting pixels, and was not suited for a moving camera due to the potential that the crop size may be too small or big to correctly crop when the camera was near or far to an interesting object. C2F-ARM does not suffer this, as all cameras are voxelised to a canonical world frame.

There are a number of areas for improvement. Currently, only raw RGB and point-cloud data are stored in the voxels, but we hypothesise that instead voxelising pixel features from a small 2D convolutional network could allow for more expressive voxel values, especially when dealing with small resolutions or a small number of coarse-to-fine Q-attention layers. Another weakness is that we are restricted to keeping the initial voxel resolution (at Q-attention depth $0$) to be reasonably small. This is not an issue when considering manipulation on a fixed table, but becomes an issue when considering mobile manipulation, where the resolution at depth $0$ may have to become very large to accommodate voxelising an entire room or house; we look to investigating solutions to this in future work. Much like ARM~\cite{james2021attention}, the control agent uses path planning and on-line trajectory generation, but will undoubtedly require improvement for achieving tasks that have dynamic environments (e.g. moving target objects, moving obstacles, etc) or complex contact dynamics (e.g. peg-in-hole). We are also keen to see if additional performance can be had by learning a fully-continuous residual function on top of the output of the coarse-to-fine network to further refine the output pose; however, we hypothesise that tasks that require grater fine-grained control will be needed for evaluation. The work that we are most excited about is exploring the use of this system in multi-task~\cite{kalashnikov2021mt} and few-shot~\cite{james2018task, bonardi2020learning} learning scenarios.

\section{Acknowledgements}

This work was supported by Dyson Technology Ltd.

{\small
\bibliographystyle{ieee_fullname}
\bibliography{egbib}
}

\end{document}